\documentclass[a4paper]{article}
\usepackage{INTERSPEECH2015}
\usepackage{epsfig}
\usepackage{algorithm,algorithmic}
\usepackage{amsmath,graphicx,paralist}

\title{The IBM 2015 English Conversational Telephone Speech Recognition System}
\name {George Saon, Hong-Kwang J. Kuo, Steven Rennie and Michael Picheny}
\address{IBM T. J. Watson Research Center, Yorktown Heights, NY, 10598\\
\tt{gsaon@us.ibm.com}}

\ninept
\copyrightnotice{\copyright\ IEEE 2004}
\toappear{Submitted to Interspeech 2015, please do not redistribute}

\begin{document}
\maketitle

\begin{abstract} 
We describe the latest improvements to the IBM English conversational
telephone speech recognition system. Some of the techniques that were
found beneficial are: maxout networks with annealed dropout rates;
networks with a very large number of outputs trained on 2000 hours of
data; joint modeling of partially unfolded recurrent neural networks
and convolutional nets by combining the bottleneck and output layers
and retraining the resulting model; and lastly, sophisticated language
model rescoring with exponential and neural network LMs. These
techniques result in an 8.0\% word error rate on the Switchboard part
of the Hub5-2000 evaluation test set which is 23\% relative better than
our previous best published result.
\end{abstract}


\section{Introduction}

Ever since~\cite{seide11} demonstrated the large accuracy gains from
using deep neural network acoustic models versus Gaussian mixture
models, the Switchboard corpus has become the de facto
standard experimental testbed for reporting believable and, more
importantly, reproducible results for LVCSR. We surmise that this is
because it is the largest publicly available dataset (up to 2300 hours
of training data) composed of truly conversational speech and because,
in general, techniques which result in improvements on Switchboard
tend to work well on both small and large vocabulary tasks. One can
think of LDA/STC, VTLN, FMLLR and lattice-based model and
feature-space discriminative training which were developed first on
Switchboard and then became ubiquitous as prime examples of such
techniques.

Since Switchboard is such a well-studied corpus, we thought we would
take a step back and reflect on how far we have come in terms of
speech recognition technology. To set the baseline, the human word
error rate on this task is estimated to be around
4\%~\cite{lippmann97}. Quoting~\cite{lippmann97} again, in 1995, ``a
high-performance HMM recognizer'' achieved a 43\% WER~on
Switchboard~\cite{liu95}. In 2000, Cambridge University achieved an at
the time impressive error rate of 19.3\% during the Hub5e DARPA
evaluation~\cite{hain00} which they attributed to ``careful
engineering''. At the height of technological development for
GMM-based systems, the winning IBM submission scored 15.2\% WER during
the 2004 DARPA EARS Rich Transcription evaluation~\cite{soltau05}
largely due to the Attila ASR toolkit~\cite{soltau10} and
fMPE~\cite{povey05}. Nowadays, deep neural networks have levelled the
playing field and multiple sites can easily reach 12-14\% WER using
much simpler systems~\cite{vesely13,seide14,hannun14,zhou14,maas14} as
shown in Table~\ref{comparison}.

To achieve an error rate of 8.0\% on this task is not trivial. In our
opinion, a successful recipe has to contain several ingredients. The
first and most obvious one is to train larger acoustic and language
models on more data. The second (a little less obvious) is to train
neural nets that have diverse architectures and operate on different
input representations so that we get accuracy gains from both feature
and model combination. Third, extra ``spice'' such as networks with
maxout nonlinearities and exponential and NN language models were also
found to significantly lower the error rate of our system. Last but
not least, it is our experience that having a strong GMM-HMM baseline
system~\cite{soltau10,soltau14} which provides high-quality alignments
used for the various speaker adaptation techniques and for DNN
cross-entropy training helps.

The paper is organized as follows: in section~\ref{gp} we describe the
processing steps that are common across all models, in
section~\ref{improvements} we present a set of system improvements,
and in section~\ref{conclusion} we summarize our findings and ponder
future opportunities for improvement.

\section{General processing}
\label{gp}

Here we describe the common processing steps for all the models
detailed in this paper. In particular, we discuss front-end
processing, speaker adaptation and neural network training specifics which are largely similar to~\cite{saon13,soltau14}.

\subsection{Training and test data}

The training data consists of 1975 hours of segmented audio from English telephone
conversations between two strangers on a preassigned topic and is
divided as follows: 262 hours from the Switchboard
1 data collection, 1698 hours from the Fisher data
collection and 15 hours of CallHome audio. The test set is
the Hub5 2000 evaluation set and contains two parts: 2.1 hours (21.4K
words, 40 speakers) of Switchboard data and 1.6 hours (21.6K words, 40
speakers) of CallHome audio. The decoding vocabulary has 30.5K words
and 32.9K pronunciations and all decodings were performed with a 4M
4-gram language model (and rescored with different LMs in
subsection~\ref{LM}).

\subsection{Feature extraction}

Speech is coded into 25 ms frames, with a frame-shift of 10 ms. Each
frame is represented by a feature vector of 13 VTL-warped perceptual
linear prediction (PLP) cepstral coefficients which are mean and
variance normalized per conversation side. Every 9 consecutive
cepstral frames are spliced together and projected down to 40
dimensions using LDA. The range of this transformation is further
diagonalized by means of a global semi-tied covariance
transform. Next, the LDA features are transformed with one
feature-space MLLR (FMLLR) transform per conversation side at both
training and test time. Convolutional nets are trained on VTL-warped
logmel features augmented with first and second temporal
derivatives. The Mel filterbank has 40 filters and the input to the
CNNs are blocks of 11 consecutive 40$\times$3-dimensional frames (as
described in~\cite{soltau14}).

In addition to VTLN and FMLLR, DNNs are adapted to the speaker by
appending 100-dimensional i-vectors to every block of 11 FMLLR frames
as described in~\cite{saon13}. The i-vectors are extracted using a
universal background model given by a GMM with 2048 diagonal
covariance mixture components which was trained with maximum
likelihood on the speaker-adapted features. The i-vectors are
extracted once per conversation side.

\subsection{Neural network training} 
All models have sigmoid hidden layers and softmax output layers
(except for the models from subsection~\ref{maxout}) and are trained
with 10-15 epochs of SGD on frame-randomized minibatches of 250 frames
and a cross-entropy criterion. The targets correspond to the
context-dependent HMM states obtained by aligning the audio with a
GMM-HMM system with 300K Gaussians trained with maximum likelihood on
the FMLLR features. The same alignments are mapped to the leaves of various
phonetic decision trees which differ in phone context size
($\pm$2 or $\pm$3) and number of leaves (16K, 32K and
64K). Prior to CE training, the networks are initialized with
layerwise discriminative pretraining as suggested
in~\cite{seide11}. Additionally, we applied 20-30 iterations of
hessian-free sequence discriminative training (ST) by using the state-based
minimum Bayes risk (MBR) objective function as described
in~\cite{bedk12}. The trained networks are used directly in a
hybrid decoding scenario by subtracting the logarithm of the HMM state
priors from the log of the DNN output scores.

\section{System improvements}
\label{improvements}

In this section we discuss specific improvements related to acoustic
and language modeling. More concretely, we describe the following
techniques: maxout models with annealed dropout
(subsection~\ref{maxout}); training DNNs, CNNs and RNNs with a very
large number of outputs (subsection~\ref{vlol}); improved joint
training of convolutional and non-convolutional nets
(subsection~\ref{joint}); and language model rescoring with
exponential and neural network LMs (subsection~\ref{LM}).

\subsection{Maxout networks with annealed dropout}
\label{maxout}
Maxout networks \cite{goodfellow13} generalize rectified linear
(ReLU, $\max[0,a]$) units, employing non-linearities of the form:
\begin{equation}
s_j = \max_{i\in C(j)} a_i
\end{equation}
~~\\
where the activations $a_i = w^T_{i}x +b_i$ are based, as usual, on
inner products, and the sets of activations $\{C(j)\}$ utilized by
different hidden units are typically disjoint. Maxout networks are
conditionally linear and so avoid the vanishing gradient problem, and
are well suited for the dropout training
procedure~\cite{srivastava14}, which for a linear model, trains an
exponentially sized model ensemble ($2^D$ models for input dimension
D), whose geometric average can be computed by simply renormalizing at
test time.

Maxout networks for ASR have recently been investigated by several
researchers, and found to produce significant gains when training data
is limited~\cite{zhang14}, but negligible gains in our personal
experience when the amount of training data exceeds approximately 100
hours. However, recently we showed that by annealing the dropout rate
over the course of training, Maxout networks can produce substantial
gains, even in big data scenarios~\cite{rennie14}. The annealing
procedure effectively initializes the ensemble of models being learned
at a given iteration with an ensemble of models with lower mean and
higher variance in the number of active units. This stochastic
regularization procedure retains the benefits of the standard dropout
training procedure (a strong exploration-phase; a preference for
population-based predictions) without compromising the capacity of the
network being learned.

Table~\ref{maxout-tab} compares the performance of our annealed
dropout Maxout networks (Maxout-AD) to corresponding sigmoid-based
DNNs and CNNs from~\cite{soltau14} learned using our standard training
procedure, using only the SWB-1 training data (262 hours). All Maxout
networks utilize 2 filters per hidden unit, and the same number of
layers and roughly the same number of parameters per layer as the
sigmoid-based DNN/CNN counterparts. Parameter equalization is achieved
by having a factor of $\sqrt{2}$ more neurons per hidden layer for the
maxout nets since the maxout operation reduces the number of outputs
by a factor of 2. Note that ReLU networks, in our experience, perform
on-par with sigmoid-based DNNs in this data regime. Maxout networks
trained with AD (Maxout-AD), on the other hand, show a clear advantage
over our traditional networks. Also, note that the convolutional layers of the Maxout-AD CNN have only 128 and 256 feature map outputs, whereas those of the sigmoid CNN has 512/512 outputs. Training of the Maxout-AD CNN with a 512/512 filter configuration is in progress.

\begin{table}[htpb!]
\begin{center}
\begin{tabular}{|c|c|c|} \hline
Model   & \multicolumn{2}{|c|}{WER SWB (ST)}\\ \cline{2-3}
        & sigmoid & Maxout-AD\\ \hline
DNN     & 11.9     & 11.0     \\ \hline
CNN     & 11.8     & 11.6     \\ \hline
DNN+CNN & 10.5     & 10.2     \\ \hline
\end{tabular}
\end{center}
\caption{\label{maxout-tab}
Word error rates of sigmoid vs. Maxout networks trained with
annealed dropout (Maxout-AD) for ST CNNs, DNNs and score fusion on Hub5'00 SWB. Note that all networks are trained only on the SWB-1 data (262 hours). }
\end{table}

\subsection{Networks with very large output layers}
\label{vlol}

When training on 2000 hours of data, we found it beneficial to
increase the number of context-dependent HMM output targets to values
that are far larger than commonly reported. To keep the computation
and the number of parameters in check, we also had to use a bottleneck
layer before the output layer~\cite{sainath13} with typically 512
neurons. Back in the days when we were training GMM-based acoustic
models, we did not notice accuracy improvements when using more than,
say, 10000 HMM states~\cite{soltau05}. We conjecture that this is
because GMMs are a distributed model and require more data for each
state to reliably estimate the mixture components, whereas the DNN
output layer is shared between states. This allows DNNs to have a much
richer target space. Additionally, we experimented with growing
acoustic decision trees where the phonetic context is increased to
heptaphones ($\pm$3 phones within words and $\pm$2 phones across
words). This was a distinct feature of our EARS RT'04 evaluation
system which made a significant difference~\cite{soltau05}. The effect of
varying the number of outputs and phonetic context is shown in
Table~\ref{vlol-tab} for DNNs with 5 hidden layers (4 with 2048 units
and 1 with 512 units) trained with 15 passes of cross-entropy on 2000 hours.

\begin{table}[htpb!]
\begin{center}
\begin{tabular}{|c|c|c|} \hline
Nb. outputs  & Phonetic ctx. & WER SWB (CE)\\ \hline 
16000        & $\pm$2        & 12.0 \\ 
16000        & $\pm$3        & 11.8 \\ \hline
32000        & $\pm$2        & 11.7 \\ \hline
64000        & $\pm$2        & 11.9 \\ \hline
\end{tabular}
\end{center}
\caption{\label{vlol-tab}
Comparison of word error rates for CE-trained DNNs with different number of outputs and phonetic context size on Hub5'00 SWB.}
\end{table}

Based on these results, a compromise was struck by choosing models
with 32K outputs and pentaphone acoustic context in all subsequent
experiments. We have trained three types of models that differ in
functionality and input features: 
\begin{itemize} 
\item Regular DNNs that operate on 11 spliced 40-dimensional FMLLR frames and 100-dimensional
i-vectors. These models have 5 hidden sigmoid layers (4 with 2048
units and 1 with 512 units) and their architecture is shown on the
left side of Figure~\ref{fig1}.  

\item Convolutional neural networks with two convolutional layers with
128 and 256 filters respectively. The CNNs operate on blocks of 11
consecutive VTL-warped 40-dimensional logmel frames augmented with
first and second derivatives with 9$\times$9 convolution windows. The convolution and pooling layer configuration is taken from~\cite{sainath13c} and the architecture is also shown on the left side of Figure~\ref{fig1}.

\item Partially unfolded recurrent neural networks~\cite{saon14} which operate on a sliding window of 6 40-dimensional FMLLR frames (from $t\ldots t+5$) and 100-dimensional i-vectors. The 6-frame window slides backwards in time from $t$ to $t-5$ (so that the RNN and the DNN have exactly the same input). The first hidden layer is recurrent and is followed by 4 non-recurrent hidden layers (3 with 2048 neurons and 1 with 512 neurons) and one output layer with 32000 softmax units.
\end{itemize}

All nets are trained with 10-15 passes of cross-entropy on 2000 hours
of audio and 30 iterations of sequence discriminative training using
Hessian-free optimization~\cite{bedk12}. The performance of the
individual networks as well as their score fusion combination is shown
in Table~\ref{net-tab} on the Hub5'00 test set (SWB and CallHome
parts). For score fusion, we decode with a frame-level sum of the
outputs of the nets prior to the softmax with uniform weights.

\begin{table}[htpb!]
\begin{center}
\begin{tabular}{|c|c|c|c|c|} \hline
Model   & \multicolumn{2}{|c|}{WER SWB} & \multicolumn{2}{|c|}{WER CH}\\ \cline{2-5}
        & CE     & ST     & CE     & ST     \\ \hline 
CNN     & 12.6 & 10.4 & 18.4 & 17.9 \\ \hline
DNN     & 11.7 & 10.3 & 18.5 & 17.0 \\ \hline
RNN     & 11.5 & 9.9  & 17.7 & 16.3 \\ \hline
DNN+CNN & 11.3 & 9.6  & 17.4 & 16.3 \\ \hline
RNN+CNN & 11.2 & 9.4  & 17.0 & 16.1 \\ \hline
DNN+RNN+CNN & 11.1 & 9.4 & 17.1 & 15.9 \\ \hline
\end{tabular}
\end{center}
\caption{\label{net-tab}
Comparison of word error rates for CE and ST CNN, DNN, RNN and various score fusions on Hub5'00.}
\end{table}

\subsection{Improved joint training of recurrent and convolutional nets}
\label{joint}

In~\cite{soltau14}, we proposed a method for jointly modeling and
training a CNN and a DNN. The crux of the method is to have the first
layers be network specific (convolution and pooling for CNN operating
on spectral features and input layer for DNN operating on PLP-based
and i-vector features) and the remaining layers be shared. The outputs
of the network-specific layers are merged into one common hidden layer
followed by additional (common) hidden layers and one output
layer. This graph structure for the joint network extends the standard
linear sequence of layers for DNNs (or CNNs). By using this
architecture, we reported a 12\% relative gain on a Switchboard 300
hours setup over the best single model (from 11.8\% for the CNN to
10.4\% for the joint CNN/DNN). We also showed that performing score
fusion of a CNN and a DNN trained separately achieves a similar WER of
10.5\%. Hence, the main benefit of the joint model in~\cite{soltau14} over
the score fusion approach is the shared computation for the common
hidden and output layers which is considerably faster than having to
do two separate forward passes.

\begin{figure}[htpb!]
\centerline{\includegraphics[scale=0.6]{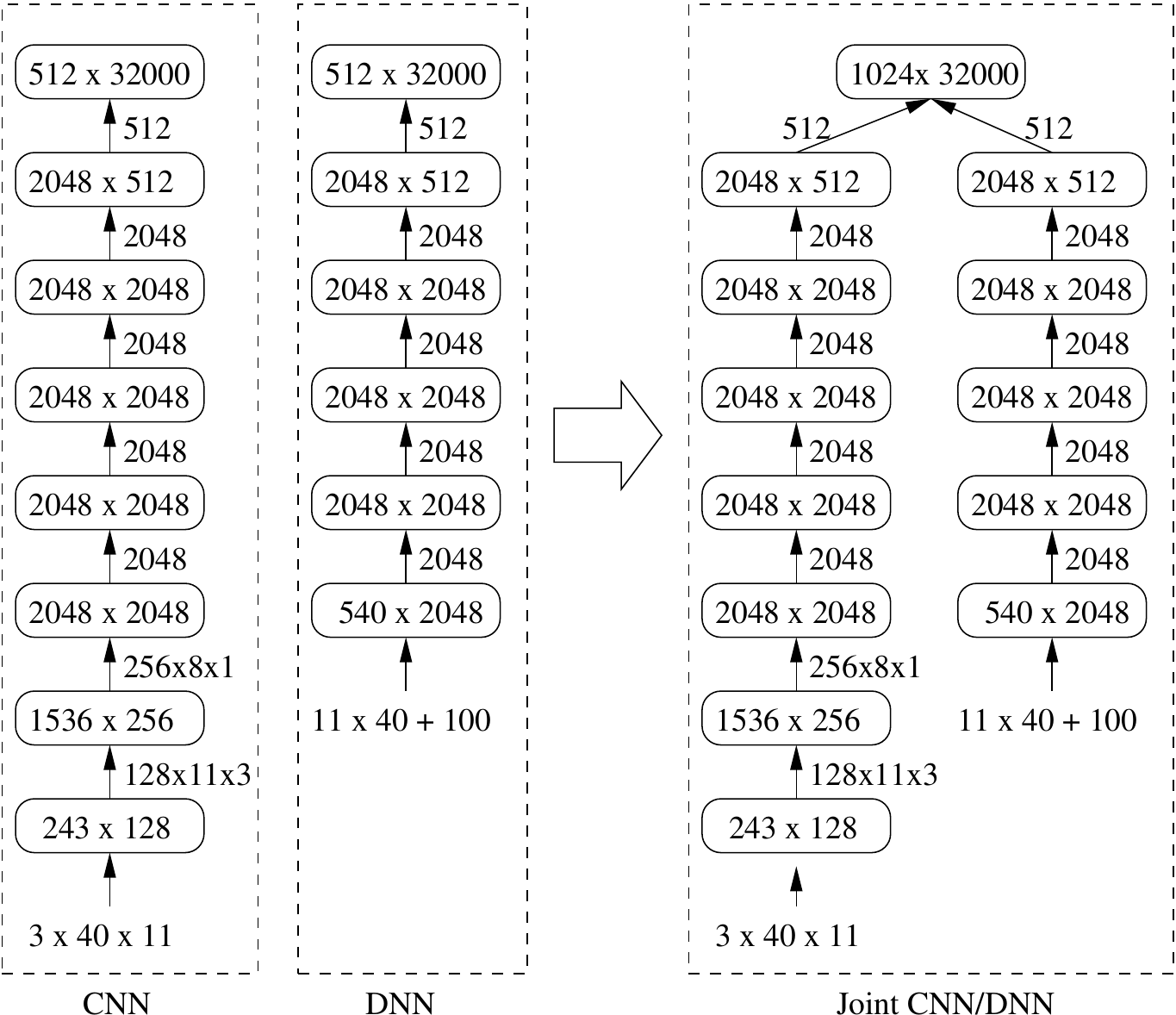}}
\caption{\label{fig1} Forming a joint CNN/DNN model out of separately trained networks by fusing the bottleneck and output layers.}
\end{figure}

A different approach that we are advocating here is to initialize the
joint model such that it is equivalent to the score fusion of the
separate models. The reasoning behind this is that, after retraining,
the objective function for sequence discriminative training can only
improve (or, at worst, remain the same). For the case of log-linear
score combination of multiple neural networks with the same number
(and type) of outputs, this initialization is done by concatenating
the individual weight matrices between the bottleneck and output
layers and by dividing the resulting matrix by the number of models
(assuming uniform weights). An example of a joint CNN/DNN model
initialized in such a way is illustrated in Figure~\ref{fig1}. For
convenience, we have indicated the sizes of the weight matrices in the
oval boxes and the dimensionality of the layers is attached to the
arrows.

We have experimented with jointly training the unfolded RNN and the CNN
from subsection~\ref{vlol}. Two experimental scenarios were
considered. The first is where the joint model was initialized with
the fusion of the cross-entropy trained RNN and CNN whereas the
second uses ST models as the starting point. For both scenarios we
generate numerator and denominator lattices with the initial joint
model and optimize the lattice-based MBR loss using distributed
hessian-free training~\cite{bedk12}. In Table~\ref{joint-tab} we
compare the WERs for several systems on the Hub5'00 test set (SWB and
CallHome parts).

\begin{table}[htpb!]
\begin{center}
\begin{tabular}{|l|c|c|} \hline
RNN/CNN combination & WER SWB & WER CH\\ \hline 
score fusion of CE models       & 11.2  & 17.0 \\ \hline
score fusion of ST models       & 9.4   & 16.1\\ \hline
joint model from CE models (ST) & 9.3   & 15.6\\ \hline
joint model from ST models (ST) & 9.4   & 15.7\\ \hline
\end{tabular}
\end{center}
\caption{\label{joint-tab}
Comparison of word error rates for CE and sequence trained unfolded RNN and DNN with score fusion and joint modeling on Hub5'00. The WERs for the joint models are after sequence training.}
\end{table}

We observe that joint modeling and sequence discriminative retraining
helps by 0.5\% on the CallHome part and only 0.1\% on SWB over score
fusion of the ST models. Also, the performance of the joint model
after sequence training appears to be slightly better for the
initialization from CE models (we expected it to be the other way
around).

\subsection{Language model}
\label{LM}

In experiments comparing acoustic models reported in previous
sections, we used a baseline legacy language model that had been used
for previous publications: a 4M 4-gram language model with a
vocabulary of 30.5K words.  While keeping the vocabulary the same, we
rebuilt the LM using publicly available (e.g. LDC) training data,
including Switchboard, Fisher, Gigaword, and Broadcast News and
Conversations.  The most relevant data are the transcripts of the 1975
hour audio data used for training the acoustic model, consisting of
about 24M words.

To build the new n-gram language model, we trained a 4-gram model with
modified Kneser-Ney smoothing~\cite{chen99} for each corpus, and then
linearly interpolated the component models with weights chosen to
optimize perplexity on a held-out set.  Then we applied entropy
pruning~\cite{stolcke98}, resulting in a single 4-gram LM consisting
of 37M n-grams.  This new n-gram LM was used in combination with our
best acoustic model to decode and generate word lattices for further
LM rescoring experiments.  The first two lines of Table~\ref{lm-tab}
show the improvement using this larger n-gram LM trained on more data.
The WER improved by 0.5\% for SWB and 0.3\% for CallHome.  Part of
this improvement (0.1-0.2\%) was due to also using a larger beam for
decoding.

\begin{table}[htpb!]
\begin{center}
\begin{tabular}{|l|c|c|} \hline
LM                        & WER SWB & WER CH \\ \hline 
Baseline 4M 4-gram        & 9.3     & 15.6   \\ \hline
37M 4-gram (n-gram)       & 8.8     & 15.3   \\ \hline
n-gram + model M          & 8.4     & 14.3   \\ \hline
n-gram + model M + NNLM   & 8.0     & 14.1   \\ \hline 
\end{tabular}
\end{center}
\caption{\label{lm-tab}
Comparison of word error rates for different language models.}
\end{table}

For LM rescoring, we used two types of LMs: model M, a class-based
exponential model~\cite{chen09} and neural network LM
(NNLM)~\cite{Bengio03,Emami06,Schwenk07,emamiasru07}.  We built a
model M LM on each corpus and interpolated the models, together with
the 37M n-gram LM.  As shown in Table~\ref{lm-tab}, using model M
results in an improvement of 0.4\% on SWB and 1.0\% on CallHome.  

We built two NNLMs for interpolation. One was trained on just the most
relevant data: the 24M word corpus (Switchboard/Fisher/CallHome
acoustic transcripts).  Another was trained on a 560M word subset of
the LM training data: in order to speed up training for this larger
set, we employed a hierarchical NNLM
approximation~\cite{Emami06,kuo2012large}.  Table~\ref{lm-tab} shows
that, compared with the n-gram LM baseline, interpolating NNLM to
model M and n-gram LM results in an improvement of 0.8\% on SWB (8.8\%
to 8.0\%) and 1.2\% on CallHome (15.3\% to 14.1\%).


Lastly, in Table~\ref{comparison} we compare our results with those
obtained by various other systems from the literature. For clarity, we
also specify the type of training data that was used for acoustic
modeling in each case.

\begin{table}[htpb!]
\begin{center}
\begin{tabular}{|l|c|c|c|} \hline
System                          & AM training data & SWB & CH\\ \hline 
Vesely et al.~\cite{vesely13}   & SWB           & 12.6  & 24.1 \\ \hline
Seide et al.~\cite{seide14}     & SWB+Fisher+other    & 13.1  & --   \\ \hline
Hannun et al.~\cite{hannun14}   & SWB+Fisher    & 12.6  & 19.3 \\ \hline
Zhou et al.~\cite{zhou14}       & SWB           & 14.2  & --   \\ \hline
Maas et al.~\cite{maas14}       & SWB           & 14.3  & 26.0 \\ \hline
Maas et al.~\cite{maas14}       & SWB+Fisher    & 15.0  & 23.0 \\ \hline
Soltau et al.~\cite{soltau14}   & SWB           & 10.4  & 19.1$^*$\\ \hline
This system                     & SWB+Fisher+CH & 8.0   & 14.1\\ \hline
\end{tabular}
\end{center}
\caption{\label{comparison}
Comparison of word error rates on Hub5'00 (SWB and CH) for existing systems ($^*$ note that 
the 19.1\% CallHome WER is not reported in~\cite{soltau14}).}
\end{table}

\section{Discussion}
\label{conclusion} 
We have presented a set of improvements to our English Switchboard
system that lowered the error rate substantially compared to our
previous best result~\cite{soltau14}. In decreasing order of
importance these are: rescoring with strong language models trained on
diverse data sources; joint training of an RNN and a CNN with 32000 outputs on
2000 hours of audio and maxout networks with annealed dropout. We
expect additional accuracy gains by training the maxout nets and
larger CNNs with a 512/512 filter configuration on all the data.

Extrapolating from historical trends, we believe that human accuracy
on this task can be reached within the next decade. We think that the way to
get there will most likely involve an increase of several orders of
magnitude in training data and the use of more sophisticated neural
network architectures that tightly integrate multiple knowledge
sources (acoustics, language, pragmatics, etc.).

\section{Ackowledgment} 
The authors wish to thank the following present and former IBM colleagues:
H.~Soltau, D.~Povey, S.~Chen, A.~Emami, V.~Goel, B.~Kingsbury,
L.~Mangu, B.~Ramabhadran, T.~Sainath and G.~Zweig for significant
contributions to the Switchboard system.

\bibliographystyle{IEEEtran}
\bibliography{inter2015}

\begin{thebibliography}{10}
\providecommand{\url}[1]{#1}
\csname url@samestyle\endcsname
\providecommand{\newblock}{\relax}
\providecommand{\bibinfo}[2]{#2}
\providecommand{\BIBentrySTDinterwordspacing}{\spaceskip=0pt\relax}
\providecommand{\BIBentryALTinterwordstretchfactor}{4}
\providecommand{\BIBentryALTinterwordspacing}{\spaceskip=\fontdimen2\font plus
\BIBentryALTinterwordstretchfactor\fontdimen3\font minus
  \fontdimen4\font\relax}
\providecommand{\BIBforeignlanguage}[2]{{%
\expandafter\ifx\csname l@#1\endcsname\relax
\typeout{** WARNING: IEEEtran.bst: No hyphenation pattern has been}%
\typeout{** loaded for the language `#1'. Using the pattern for}%
\typeout{** the default language instead.}%
\else
\language=\csname l@#1\endcsname
\fi
#2}}
\providecommand{\BIBdecl}{\relax}
\BIBdecl

\bibitem{seide11}
F.~Seide, G.~Li, X.~Chien, and D.~Yu, ``Feature engineering in
  context-dependent deep neural networks for conversational speech
  transcription,'' in \emph{Proc. ASRU}, 2011.

\bibitem{lippmann97}
R.~P. Lippmann, ``Speech recognition by machines and humans,'' \emph{Speech
  communication}, vol.~22, no.~1, pp. 1--15, 1997.

\bibitem{liu95}
F.~Liu, M.~Monkowski, M.~Novak, M.~Padmanabhan, M.~Picheny, and P.~Rao,
  ``Performance of the {IBM} {LVCSR} system on the {S}witchboard corpus,'' in
  \emph{Proceedings of Speech Research Symposium}, 1995, p. 189.

\bibitem{hain00}
T.~Hain, P.~Woodland, G.~Evermann, and D.~Povey, ``The {CU-HTK} march 2000
  {HUB5E} transcription system,'' in \emph{Proc. Speech Transcription
  Workshop}, vol.~1.\hskip 1em plus 0.5em minus 0.4em\relax Baltimore, 2000.

\bibitem{soltau05}
H.~Soltau, B.~Kingsbury, L.~Mangu, D.~Povey, G.~Saon, and G.~Zweig, ``The {IBM}
  2004 conversational telephony system for rich transcription.'' in
  \emph{Acoustics, Speech and Signal Processing (ICASSP), 2005 IEEE
  International Conference on}, 2005, pp. 205--208.

\bibitem{soltau10}
H.~Soltau, G.~Saon, and B.~Kingsbury, ``The {IBM} {A}ttila speech recognition
  toolkit,'' in \emph{Proc. of IEEE Workshop on Spoken Language Technology
  (SLT)}, 2010, pp. 97--102.

\bibitem{povey05}
D.~Povey, B.~Kingsbury, L.~Mangu, G.~Saon, H.~Soltau, and G.~Zweig, ``{fMPE}:
  {D}iscriminatively trained features for speech recognition,'' in \emph{Proc.
  of ICASSP}, 2005, pp. 961--964.

\bibitem{vesely13}
K.~Vesely, A.~Ghoshal, L.~Burget, and D.~Povey, ``Sequence-discriminative
  training of deep neural networks,'' in \emph{Proc. Interspeech}, 2013.

\bibitem{seide14}
F.~Seide, H.~Fu, J.~Droppo, G.~Li, and D.~Yu, ``1-bit stochastic gradient
  descent and its application to data-parallel distributed training of speech
  dnns,'' in \emph{Fifteenth Annual Conference of the International Speech
  Communication Association}, 2014.

\bibitem{hannun14}
A.~Hannun, C.~Case, J.~Casper, B.~Catanzaro, G.~Diamos, E.~Elsen, R.~Prenger,
  S.~Satheesh, S.~Sengupta, A.~Coates \emph{et~al.}, ``Deepspeech: Scaling up
  end-to-end speech recognition,'' \emph{arXiv preprint arXiv:1412.5567}, 2014.

\bibitem{zhou14}
P.~Zhou, L.~Dai, and H.~Jiang, ``Sequence training of multiple deep neural
  networks for better performance and faster training speed,'' in
  \emph{Acoustics, Speech and Signal Processing (ICASSP), 2014 IEEE
  International Conference on}.\hskip 1em plus 0.5em minus 0.4em\relax IEEE,
  2014, pp. 5627--5631.

\bibitem{maas14}
A.~L. Maas, A.~Y. Hannun, C.~T. Lengerich, P.~Qi, D.~Jurafsky, and A.~Y. Ng,
  ``Increasing deep neural network acoustic model size for large vocabulary
  continuous speech recognition,'' \emph{arXiv preprint arXiv:1406.7806}, 2014.

\bibitem{soltau14}
H.~Soltau, G.~Saon, and T.~N. Sainath, ``Joint training of convolutional and
  non-convolutional neural networks,'' \emph{to Proc. ICASSP}, 2014.

\bibitem{saon13}
G.~Saon, H.~Soltau, D.~Nahamoo, and M.~Picheny, ``Speaker adaptation of neural
  network acoustic models using i-vectors,'' in \emph{Proc. ASRU}, 2013.

\bibitem{bedk12}
B.~Kingsbury, T.~Sainath, and H.~Soltau, ``Scalable minimum {B}ayes risk
  training of deep neural network acoustic models using distributed
  {H}essian-free optimization,'' in \emph{Proc. Interspeech}, 2012.

\bibitem{goodfellow13}
I.~J. Goodfellow, D.~Warde-Farley, M.~Mirza, A.~Courville, and Y.~Bengio,
  ``Maxout networks,'' \emph{arXiv preprint arXiv:1302.4389}, 2013.

\bibitem{srivastava14}
N.~Srivastava, G.~Hinton, A.~Krizhevsky, I.~Sutskever, and R.~Salakhutdinov,
  ``Dropout: A simple way to prevent neural networks from overfitting,''
  \emph{The Journal of Machine Learning Research}, vol.~15, no.~1, pp.
  1929--1958, 2014.

\bibitem{zhang14}
X.~Zhang, J.~Trmal, D.~Povey, and S.~Khudanpur, ``Improving deep neural network
  acoustic models using generalized maxout networks,'' in \emph{Acoustics,
  Speech and Signal Processing (ICASSP), 2014 IEEE International Conference
  on}.\hskip 1em plus 0.5em minus 0.4em\relax IEEE, 2014, pp. 215--219.

\bibitem{rennie14}
S.~Rennie, V.~Goel, and S.~Thomas, ``Annealed dropout training of deep
  networks,'' in \emph{Spoken Language Technology (SLT), IEEE Workshop on.
  IEEE}, 2014.

\bibitem{sainath13}
T.~Sainath, B.~Kingsbury, V.~Sindhwani, E.~Arisoy, and B.~Ramabhadran,
  ``{Low-rank matrix factorization for deep neural network training with
  high-dimensional output targets},'' in \emph{Proc. of ICASSP}, 2013.

\bibitem{sainath13c}
T.~N. Sainath, A.-r. Mohamed, B.~Kingsbury, and B.~Ramabhadran, ``Deep
  convolutional neural networks for {LVCSR},'' in \emph{Acoustics, Speech and
  Signal Processing (ICASSP), 2013 IEEE International Conference on}.\hskip 1em
  plus 0.5em minus 0.4em\relax IEEE, 2013, pp. 8614--8618.

\bibitem{saon14}
G.~Saon, H.~Soltau, A.~Emami, and M.~Picheny, ``Unfolded recurrent neural
  networks for speech recognition,'' in \emph{Fifteenth Annual Conference of
  the International Speech Communication Association}, 2014.

\bibitem{chen99}
S.~F. Chen and J.~Goodman, ``An empirical study of smoothing techniques for
  language modeling,'' \emph{Computer Speech \& Language}, vol.~13, no.~4, pp.
  359--393, 1999.

\bibitem{stolcke98}
A.~Stolcke, ``Entropy-based pruning of backoff language models,'' in
  \emph{Proc. DARPA Broadcast News Transcription and Understanding Workshop},
  1998, pp. 270--274.

\bibitem{chen09}
S.~F. Chen, ``Shrinking exponential language models,'' in \emph{Proc.
  NAACL-HLT}, 2009, pp. 468--476.

\bibitem{Bengio03}
Y.~Bengio, R.~Ducharme, P.~Vincent, and C.~Jauvin, ``A neural probabilistic
  language model,'' \emph{Journal of Machine Learning Research}, vol.~3, pp.
  1137--1155, 2003.

\bibitem{Emami06}
A.~Emami, ``A neural syntactic language model,'' Ph.D. dissertation, Johns
  Hopkins University, Baltimore, MD, USA, 2006.

\bibitem{Schwenk07}
H.~Schwenk, ``Continuous space language models,'' \emph{Computer Speech \&
  Language}, vol.~21, no.~3, pp. 492--518, 2007.

\bibitem{emamiasru07}
A.~Emami and L.~Mangu, ``Empirical study of neural network language models for
  {A}rabic speech recognition,'' in \emph{Proc. ASRU}, 2007, pp. 147--152.

\bibitem{kuo2012large}
H.-K.~J. Kuo, E.~Ar{\i}soy, A.~Emami, and P.~Vozila, ``Large scale hierarchical
  neural network language models,'' in \emph{Proc. Interspeech}, 2012.

\end{thebibliography}

\end{document}